\documentclass[letterpaper, 10 pt, conference]{ieeeconf}
\makeatletter
\pdfoutput=1
\let\NAT@parse\undefined
\makeatother
\IEEEoverridecommandlockouts 
\usepackage[sort&compress,numbers]{natbib}
\usepackage{times}  
\usepackage{helvet}  
\usepackage{courier}  
\usepackage[hyphens]{url}  
\usepackage{graphicx} 
\usepackage{caption} % [font=small]
\usepackage[dvipsnames]{xcolor}
\usepackage{algorithm}
\usepackage{algorithmic}
\usepackage{subcaption}
\usepackage{colortbl}
\usepackage{booktabs}
\usepackage{fix-cm}
\usepackage{newfloat}
\usepackage{listings}

\newfloat{listing}{tb}{lst}{}
\floatname{listing}{Listing}

\usepackage{amsmath}
\usepackage{amssymb}
\usepackage{tabularx}
\usepackage{arydshln}
\usepackage{multirow}
\usepackage{lmodern}
\usepackage{anyfontsize} 
\usepackage{bm}
\usepackage[shrink=190,letterspace=500]{microtype}
\microtypecontext{spacing=nonfrench}

\def\secref#1{\hyperref[#1]{Section~\ref{#1}}}
\def\figref#1{\hyperref[#1]{Fig.~\ref{#1}}}
\def\tabref#1{\hyperref[#1]{Tab.~\ref{#1}}}
\def\eqref#1{\hyperref[#1]{Eq.~\ref{#1}}}

\newcommand{\xvec}{\mathbf{x}}
\newcommand{\nbf}[1]{{\noindent \textbf{#1.}}}

\usepackage{hyperref}
\hypersetup{
     colorlinks=true,
     linkcolor=blue,
     filecolor=magenta,      
     urlcolor=cyan,
     citecolor=blue,
     pdftitle={CRiTIC},
     }
\title{\LARGE \bf Curb Your Attention: Causal Attention Gating for Robust Trajectory Prediction in Autonomous Driving}

\author {
    Ehsan Ahmadi\textsuperscript{\rm 1,2},
    Ray Mercurius\textsuperscript{\rm 3},
    Soheil Alizadeh\textsuperscript{\rm 2},
    Kasra Rezaee\textsuperscript{\rm 2},
    Amir Rasouli\textsuperscript{\rm 2}
    \thanks{$^{1}$ University of Alberta. eahmadi@ualberta.ca $^{2}$ Noah's Ark Laboratory, Huawei Technologies Canada. first.last@huawei.com. $^{3}$ Cornell University. Work done during internship at Huawei Technologies Canada.}%
}

\begin{document}

\maketitle
\thispagestyle{plain}
\pagestyle{plain}

\begin{abstract}
Trajectory prediction models in autonomous driving are vulnerable to perturbations from non-causal agents whose actions should not affect the ego-agent's behavior.
Such perturbations can lead to incorrect predictions of other agents' trajectories, potentially compromising the safety and efficiency of the ego-vehicle's decision-making process. Motivated by this challenge, we propose \textit{Causal tRajecTory predICtion} (CRiTIC), a novel model that utilizes a causal discovery network to identify inter-agent causal relations over a window of past time steps. To incorporate discovered causal relationships, we propose a novel \textit{Causal Attention Gating} mechanism to selectively filter information in the proposed Transformer-based architecture.

We conduct extensive experiments on two autonomous driving benchmark datasets to evaluate the robustness of our model against non-causal perturbations and its generalization capacity.
Our results indicate that the robustness of predictions can be improved by up to $\mathbf{54\%}$ without a significant detriment to prediction accuracy.
Lastly, we demonstrate the superior domain generalizability of the proposed model, which achieves up to $\mathbf{29\%}$ improvement in cross-domain performance. 
These results underscore the potential of our model to enhance both robustness and generalization capacity for trajectory prediction in diverse autonomous driving domains.\footnote[4]{The project page is hosted at: \url{http://ehsan-ami.github.io/critic}}
\end{abstract}

\section{Introduction}
\label{sec:intro}
Trajectory prediction is a vital part of autonomous driving (AD) research and has received increased attention in recent years, witnessing many advancements in this field \cite{multipath++, mtr, scene_transformer, wayformer}.
Given the safety-critical nature of autonomous driving systems, the robustness of prediction models to noise and distribution shifts is of utmost importance.

\begin{figure*}[tp!]
\centering
\includegraphics[width=\linewidth]{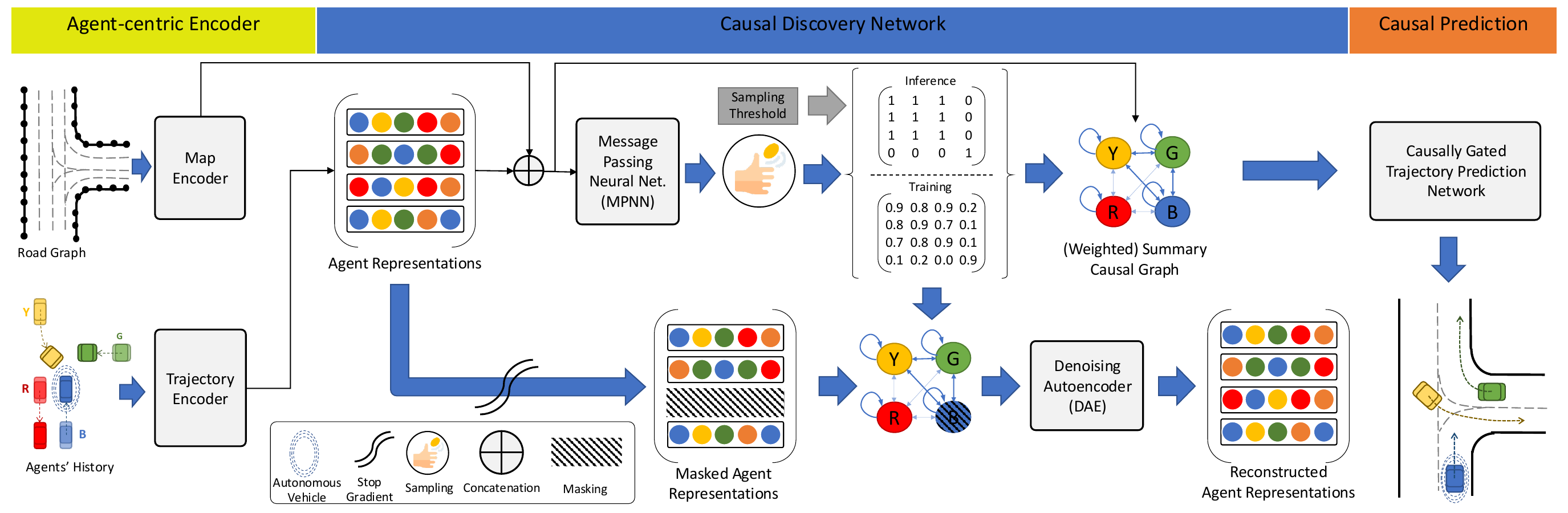}
\caption{An overview of CRiTIC. In this architecture, Causal Discovery Network receives the agent representations and generates a causality adjacency matrix. The matrix is used by a Transformer-based prediction backbone to shape the attention toward the causal agents.}
\label{fig:ctp} 
\vspace{-0.7cm}
\end{figure*}

Existing prediction models, despite achieving high accuracy, often heavily rely on \textit{spurious features} such as the behavior of non-causal agents. Such a reliance leads to safety-critical issues and limits generalization of these models to new scenarios \cite{Causal_agents}. For example, consider the scenario shown in \figref{fig:ctp}. Vehicles \textbf{Y} and \textbf{R} exhibit highly correlated trajectories until reaching the intersection. At this point, their correlation breaks as \textbf{Y} intends to turn left. Here, vehicle \textbf{B} is the autonomous vehicle whose behavior depends on the prediction results. A model exploiting the correlation between \textbf{R} and \textbf{Y} might incorrectly predict that \textbf{Y} would continue following \textbf{R}, potentially causing a collision between \textbf{Y} and \textbf{B}. As shown in \figref{fig:qual_samples}, the vulnerability of state-of-the-art models to non-causal agent removal perturbation exists in real-world scenarios (top row). This problem, however, can be remedied by effectively distinguishing between causal and non-causal relationships (bottom row). 

To suppress the impact of spurious correlations, \textit{causal representation learning} methods can be adopted to learn invariant causal latent variables \cite{causal_representation_learning}.
For instance, in the inertia and collision problems in imitation learning, causal variables are extracted from input latent space to train the motion planner \cite{Causal_im}.
The model in \cite{liu2022towards} splits the variables into invariant, spurious, and style confounder features for pedestrian motion forecasting. 
For both of the approaches the learned representations are scene-centric.
Although shown to be effective, compared to object-centric causal discovery approaches, causal disentanglement in scene-centric latent space is less interpretable and controllable (more about this in \secref{Causal Graph Discovery}).

\newcommand{\figWidthm}{0.95\linewidth}

\begin{figure}[t!]
\centering
\begin{tabular}{@{}m{3.5mm}@{} m{\figWidthm}} 
  & \fontsize{7pt}{1pt}\selectfont \;\;\;\;\;\;\;\;\;\;\;\;\;\;\;\; Original  \;\;\;\;\; \;\;\;\;\; \;\;\;\;\;\;\;\;\;\;\;\;\;\;\;\fontsize{7pt}{1pt}\selectfont Remove Noncausal \\
  \rotatebox[origin=c]{90}{\fontsize{7pt}{1pt}\selectfont CRiTIC (Ours) \;\;\;\;\;\;\;\;\;\;\;\;\;\;\;\;\;\;\;\;\;\; MTR \cite{mtr}} & \includegraphics[clip, trim=0.0cm 0.0cm 0.0cm 0.0cm, width=\figWidthm]{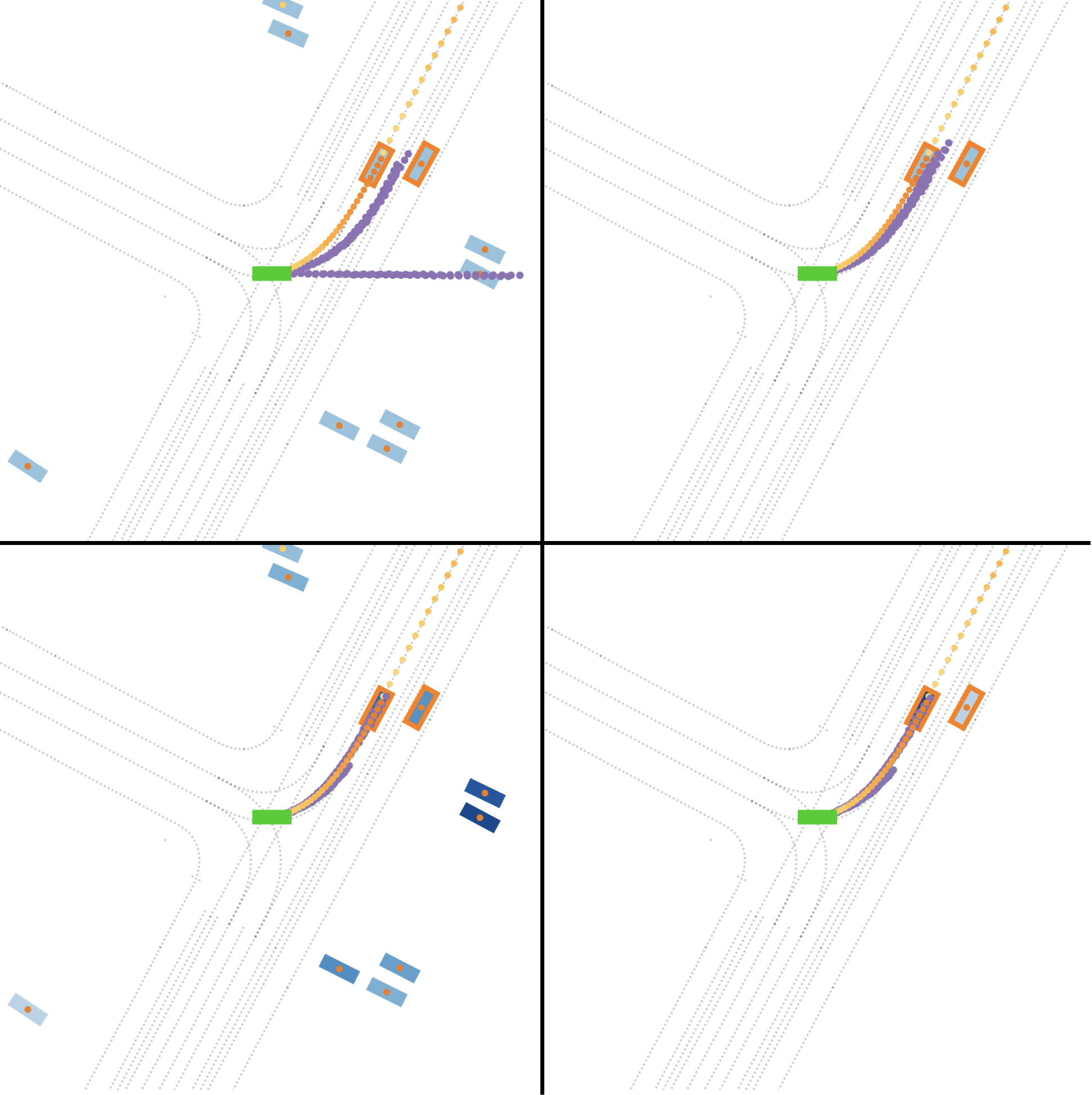}\\
\end{tabular} 
\caption{ Robustness qualitative samples.
The AV is shown in green. In CRiTIC's scene visualizations, the likelihood of a non-ego agent being causal is shown by its color saturation. 
The orange borderline indicates that the agent is labeled causal based on human labels \cite{Causal_agents}. 
The ground truth, and predictions are shown in orange, and purple colors, respectively.
Our model's performance is less affected by the intervention compared with the non-causal model.}
\label{fig:qual_samples}
\end{figure}

To this end, we propose CRiTIC, a novel object-centric causal trajectory prediction model that employs a separate \textit{Causal Discovery Network} (CDN) to identify causal inter-agent relations and predict the future trajectory conditioned on the causal graph estimated by the CDN.
The CRiTIC is designed to enhance robustness and generalizability in trajectory prediction.
For causal discovery, we create an information bottleneck (IB) in the training process to force the model to dedicate its limited attention capacity to the most relevant agents.  We also use a self-supervised \textit{graph structure learning} (GSL) auxiliary task to improve the performance of the CDN.

To integrate the resulting causal graph into the prediction module, we introduce a \textit{causal attention gating} (CAG) mechanism that suppresses information from non-causal agents in the attention sub-layers of Transformer networks, which are extensively used in trajectory prediction models and other applications. 

In summary, our \textbf{contributions} are as follows:
    \begin{itemize}
        \item We propose CRiTIC, a novel agent-centric causal model with explicit inter-agent causal relation reasoning (formulated in \secref{sec:background}) to have enhanced interpretability, robustness, generalization, and inference-time social attention controllability for trajectory prediction task (\secref{Causal Graph Discovery}).
        \item We propose a novel causal attention-gating mechanism that modulates the attention weights in our Transformer-based trajectory prediction model to only pay attention to the proposed causal agents by the CDN. This mechanism creates a dynamic information bottleneck which is effective in identifying causal relations (\secref{sec:CRiTIC}).
        \item We conduct extensive experiments, showcasing the enhanced robustness and superior generalization capacity of our model thanks to its causal reasoning power (\secref{sec:experiments}).
    \end{itemize}

\section{Related Work}
\label{sec:related}
\nbf{Graph Structure Learning} Given a set of input feature vectors, the goal of GSL is to simultaneously learn the graph structure and the node representations that are optimized for a downstream task \cite{zhu2021survey}.
Several approaches have been proposed, including neural NRI, which learns the latent structure using VAEs \cite{NRI_18}, IDGL that iteratively learns node features and graph structure using bi-level optimization \cite{IDGL}, LDS, which models edges with learned Bernoulli distribution, and uses a sampling approach to generate the graph \cite{franceschi2019learning}, and SLAPS that relies on a denoising autoencoder (DAE) in a self-supervised manner to reconstruct the node features and predict the node labels simultaneously \cite{SLAPS_lgi21}. 

In our work, we are mainly interested in the application of the GSL in causal structure discovery to infer causal relations in observational data. There are three main approaches to causal structure discovery: constraint-based \cite{entner2010causal}, score-based \cite{chickering2002optimal}, and learning-based \cite{tank2021neural}. 
The objective of these methods is to identify a shared causal graph from which all the sample data is generated.

The amortized causal discovery (ACD) method exploits the idea of \textit{Granger-causality} (see \secref{sec:background}) to identify causal graphs from purely observational data. It assumes data samples are generated by various causal graphs, while their underlying dynamics model is shared \cite{amortized_causal_discovery}.
This method benefits from a separate causal discovery module and a dynamics model which is conditioned on the causal graph generated by the former module. Hence, ACD is a suitable causal discovery approach, however, it is not designed for trajectory prediction in traffic scenarios where the model needs to process complex context information, e.g. various map elements, traffic lights, and multimodal futures. Our work builds on top of ACD's formulation for amortized causal discovery from observational data and extends it for the application of autonomous driving.

\nbf{Trajectory Prediction}
Autonomous driving systems rely heavily on accurate prediction of other road users, such as pedestrians \cite{Rasouli_ICRA_2023, li2022graph, Bae_2022_ECCV, Choi_2024_IV, rasouli2021bifold, Shi_2021_CVPR} and other vehicles \cite{Pourkeshavarz_2024_CVPR, Karim_2024_ICRA, kim2021lapred, zhou2022hivt, wang2022ltp, liang2020learning} to ensure safe and efficient navigation through dynamic environments. 
There are several components that can be the focus of the research in this area, such as representation and encoding of the input data \cite{amirloo2022latentformer, Cui_2021_ICCV, scene_transformer, Casas_2020_ECCV, Girgis_2022_ICLR, gao2020vectornet, Pourkeshavarz_2023_ICCV, Zhang_2024_IV}, multi-modality and intention prediction \cite{mfp2019, Salzmann_2020_ECCV, gilles2022gohome, multipath++, lee2022muse, zhao2021tnt,Rasouli_2024_ICRA}, and interaction modeling between the agents \cite{scene_transformer, yuan2021agentformer,clstm2018}. 

Prediction models often focus on feature correlations without considering meaningful causal relations, which are crucial for enhancing robustness and generalizability. Recent approaches have incorporated causal feature learning to address this issue. For instance, the model in \cite{liu2022towards} improves out-of-distribution generalization by categorizing latent variables into domain-invariant causal variables, domain-specific confounders, and spurious features, suppressing the latter during training to rely on causal features. Similarly, the authors of \cite{chen2021human} propose a counterfactual analysis method that mitigates environmental bias using counterfactual interventions. However, while these models utilize scene-centric representations, they lack explicit modeling of inter-agent causal relations and do not employ disentangled representations for individual agents.

To investigate the robustness of the prediction models, the authors of \cite{Causal_agents} selected three state-of-the-art trajectory models, namely MultiPath++ \cite{multipath++}, SceneTransformer \cite{scene_transformer}, and Wayformer \cite{wayformer} and  evaluated them under various perturbations applied to the data. 
The authors used human-annotated data to apply non-causal agent perturbations and showed that the performance of the aforementioned models was heavily impacted. They propose a heuristic data augmentation method to remove context vehicles during the training time. We, however, opt for an end-to-end model-centric approach that does not require heuristic function engineering.
Our model addresses the causal robustness issue by explicitly identifying the causal graph and employing it in the trajectory prediction process.

\section{Problem Formulation}
\label{sec:background}
\nbf{Trajectory Prediction}
The goal is to predict future trajectories (in $XYZ$ coordinates) of target agents $\mathbf{T} \in \mathbb{R}^{N_{\text{targ}} \times M \times T_f \times 3}$ (where $N_{\text{targ}}$ represents the number of target agents, $M=6$ and $T_f$ denote the number of modes and future time steps, respectively) given the observed agents' trajectories
$\mathbf{S}_A$ and scene context $\mathbf{S}_M$ as: $\mathbf{T} = \textup{F}(\mathbf{S}_A, \mathbf{S}_M) $.
The observed trajectories $\mathbf{S}_A \in \mathbb{R}^{N_{\text{obs}} \times T_{h} \times D_a^0}$ (where $N_{\text{obs}}$ is the number of observed agents in the scene, $T_h$ is the number of history time steps, and $D_a^0$ is the dimension of agent features) include features, such as coordinates, heading angle, velocity, dimensions, and agent type.
The scene context $\mathbf{S}_M \in \mathbb{R}^{N_m \times N_p \times D_m^0}$ represents map components (e.g., lanes, road boundaries) as $N_m$ polylines, each consisting of up to $N_p$ points, and characterized by position, component type, and dynamic states with the total feature dimension of $D_m^0$.

\nbf{Causal Graphical Models}
The Causal Graphical model describes the distribution of a set of random variables with a \textit{Directed Acyclic Graph} (DAG), which is represented as $\mathcal{G}^{1:T} = \{\mathcal{V}^{\text{1}:T}, \mathcal{E}^{1:T}\}$. Here, we assume that the node variables are based on $N$ time series $\mathcal{V}^{1:T} = \{\mathcal{V}^{1: T}_i\}_{i=1}^{N}$. The edges $\mathcal{E}^{1:T} = \{(v_i^t, v_j^t) | v_i^t, v_j^t \in \mathcal{V}^{1:T}, v_i^t \: \text{causes} \: v_j^t\}$ represent the causal relationships between the variables. Based on \textit{causal Markov assumption}, a variable $v_i^t$ is independent of its non-decendents given its parents. The \textit{Summary Causal Graph} (SCG) (referred to as a causal graph for brevity), $\mathcal{G} = \{\mathcal{V}, \mathcal{E}\}$, is a compact version of the temporal causal graph,  which captures the overall causal relationships between variables across all time steps: $\mathcal{E} = \{(v_i, v_j) | \exists t,t': (v_i^t, v_j^{t'}) \in \mathcal{E}^{1:T}\}$, $\mathcal{V}=\{v_i\}_{i=1}^{N}$.

Granger causality is a commonly used method in the domain of observational time series data \cite{granger1969investigating}, which in our case is well suited for the trajectory prediction problem. A variable $X$ is said to ``Granger-cause" variable $Y$ if past values of $X$ provide statistically significant information about future values of $Y$, beyond what is already contained in the past values of $Y$ alone. 

For causal trajectory prediction, we adopt the formulation of the amortized causal discovery (ACD) approach, which builds upon the principles of Granger causality \cite{amortized_causal_discovery}. Its key assumption is that for a set of data samples $\xvec_s$, with a varying set of summary causal graphs, $\mathcal{G}_s$, there exists a shared dynamics model $g$ that $\xvec^{t+1}_s = g(\xvec_s^{\leq t}, \mathcal{G}_s) + \bm \varepsilon_s^{t+1}$. 
Here, $\mathcal{G}_s$ represents the causal graph for sample $s$, $\xvec_s \in \mathbb{R}^{N_{\text{s}} \times T}$ denotes $N_s$ time series of length $T$, and $\varepsilon_s^{t+1}$ is the independent noise term modeling inherent aleatoric uncertainty of the data.
This formulation allows us to separate the data generation process into two components and express the process as: $\xvec^{t+1}_s \approx f_\theta(\xvec_s^{\leq t}, f_\phi(\xvec_s^{\leq t}))$, with the components:
(\textit{i}) causal discovery model $f_\phi$ ---this is the CDN module explained in \secref{Causal Graph Discovery}--- which infers the causal structure from the observed data, and (\textit{ii}) shared dynamics 
model $f_\theta$ ---this is the trajectory prediction backbone, $\mathbf{T}$, discussed in \secref{sec:bacbone}--- which predicts future states based on past observations and the inferred causal structure.

\section{Methods} %Trajectory Predictor}
\label{sec:method}

CRiTIC is composed of three modules: \textbf{(i)} AgentNet which processes each agent and its surrounding map locally (\secref{agent_net}); \textbf{(ii)} Causal Discovery Network (CDN) that constructs the causal graph based on the past interactions of the agents (\secref{Causal Graph Discovery}); and \textbf{(iii)} trajectory prediction backbone (\secref{sec:CRiTIC}), which acts as the shared dynamics model as introduced in \secref{sec:method} and predicts the future given the agent representations and the causal graph. 

\subsection{AgentNet: Generating Agent-Centric Representations}
\label{agent_net}
AgentNet is composed of two submodules: \textit{trajectory encoder} and \textit{map-encoder}.
First, the raw map and agent inputs are presented in an agent-centric approach \cite{mtr,multipath++}.
The context information generated by the map encoder is fused with the outputs of the trajectory encoder to form the contextual agent representations.
Note that in this stage the agents' states are encoded independently, as we need disentangled agent representations for explicit causal agent interaction modeling via the CDN. 

\nbf{Map Encoder}
We encode each map component using a PointNet-based encoder \cite{pointnet}, followed by a max-pooling layer, $\mathbf{S}'_M = \textup{MaxPool(PFFN}(\mathbf{S}_M))$,
where PFFN is a point-wise feed-forward network, $\mathbf{S}'_M \in \mathbb{R}^{N_m \times D_m^1}$ is the matrix of map features, and $D_m^1$ is the dimension of the map features.
To efficiently incorporate map information into agent representations, we employ a \textit{multi-context gating} (MCG) layer \cite{multipath++} to gather information from $N_m$ map elements into a concise map context vector $\mathbf{S}''_M \in \mathbb{R}^{D_m^2}$, where $\mathbf{S}''_M = \textup{MCG}(\mathbf{S}'_M)$.

\nbf{Context-aware Trajectory Encoder} The agents' states, while presented in the target agent's coordinate frame, are processed using a Pointnet-based encoder \cite{pointnet}, followed by a max-pooling and a GRU layer \cite{gru} used to encode the Kinematics of the agents.
To get intermediate temporal trajectory representations we have  $\mathbf{S}'_A = \textup{MLP} (\mathbf{S}_A)$, where $\mathbf{S}'_A \in \mathbb{R}^{N_{obs} \times T \times D_a^1}$ and $D_a^1$ is the dimension of the intermediate agent representations. Then, 
$\mathbf{S}''_A = \textup{GRU}(\textup{MaxPool} (\mathbf{S'}_A)) \oplus \mathbf{S}''_M$,
where $\oplus$ represents the concatenation operator. Here, the single map context vector $\mathbf{S}''_M$ is concatenated with each agent's representation separately. This results in $\mathbf{S}''_A \in \mathbb{R}^{N_{obs} \times D_a^2}$, the matrix of map-aware agent representations with dimension $D_a^2$.

\subsection{Causal Discovery Network (CDN)}
\label{Causal Graph Discovery}
We capture the inter-agent causal relations in the form of a directed graph, where the nodes are the agent representations and the edges indicate the causal relations among the agents.
The role of CDN is to process the set of map-aware agent representations generated by AgentNet and identify the structure of the causal summary graph. 
In other words, CDN determines whether for each pair of agents $(i,j)$, whether agent $i$ has a causal influence on agent $j$.

For trajectory prediction problem we opt for summary causal graph, as it captures all the causal relations over the span of past $T_h$ time steps. It also doesn't need to satisfy extra graph acyclicity constraint of instantaneous causal graphs, which are DAGs making the training process of CDN more tractable.
As shown in \figref{fig:ctp}, CDN is made of two major components discussed below:

\nbf{Message Passing Neural Network (MPNN)} Inspired by the expressiveness of the MPNNs \cite{gilmer2017mpnn} in processing of the dynamic relations of moving objects in \cite{NRI_18, amortized_causal_discovery}, we adopt a single layer MPNN as the core of the causal discovery network:
% mutliline format
\begin{equation}
\label{eq:mpnn}
\begin{split}
\mathbf{s}^1_i &=\gamma^{0}(\mathbf{s}^0_i),\\
\mathbf{s}^{2}_i &= \gamma^{1} \left( \mathbf{s}^{1}_i ,\sum_{j \in \mathcal{N}(i)} \phi^{1} \left( \mathbf{s}^{1}_i, \mathbf{s}^{1}_j\right) \right),\\
e_{ij} &= \left\{  \begin{matrix}
    \rho (\mathbf{W_\text{MPNN}}^T\phi^{2}(\mathbf{s}^{2}_i, \mathbf{s}^{2}_j), \lambda)& i\neq j \\
    1&  i=j
   \end{matrix}    \right.,\\
\end{split}
\end{equation}
where $\mathbf{s}^0_i \in \mathbf{S}''_A$ is an individual initial node representation in the MPNN, and it is set to representation vector of  the agent $i$ generated by AgentNet. Consequently, $\mathbf{s}^1_i$ and $\mathbf{s}^2_i$ are intermediate node representations of agent $i$. The set $\mathcal{N}(i)$ includes neighbors of node $i$ (we assume to have an initial fully-connected adjacency graph without self-loops), $\gamma^{0}, \gamma^{1}, \phi^{1}, \phi^{2}$ are two-layer MLP networks and $\mathbf{W}_{\text{MPNN}}$ is a learnable linear projection weight. 

Causal edges could be presented by binary discrete random variables which have Bernoulli distributions. However, training models with discrete variables is challenging due to the non-differentiability of the sampling operation. 
To overcome this issue, we replace the binary edge variables with a low-variance continuous relaxation of it named ``BinConcrete" \cite{gumbel_trick}, which is represented by the function $\rho$ in the above equation. 
It is defined as: $\text{BinConcrete}(\alpha,\lambda) \doteq  \text{Sigmoid}((L+\log \alpha)/\lambda)$, where $\lambda$ is the temperature hyperparameter and $L = \log U + \log (1-U)$, where $U$ is a random variable sampled from a uniform distribution. 
Finally, the weighted edges $e_{ij} \in [0,1]$ form the weighted adjacency matrix $\mathbf{A}$. 

During inference, we apply a confidence threshold value $\tau$ to obtain a discrete causal graph. This approach allows us to adjust the sparsity of the causal graph at the inference time via the threshold value. We further discuss this feature with insights from the empirical results in \secref{sec:experiments}.

\noindent \textbf{Sparsity Regularization.} A possible degenerate state for the causal discovery network is always generating a densely connected graph, or in the extreme case, a fully connected graph. Obviously, such a graph is not able to discriminate causal vs. non-causal relations.
Inspired by the sparsity of causal interactions among the agents \cite{Causal_agents}, we add an adjacency matrix sparsity regularization loss. This loss is defined as the KL divergence between the marginal probability of an edge being causal and a fixed prior binary distribution with the hyperparameter $p$ indicating the marginal probability for having a causal link. 
A small value of this parameter creates an information bottleneck by inducing a sparse causal interaction graph. The causally gated backbone prediction model, as explained later in \ref{sec:CRiTIC}, has limited information interchange bandwidth among agent representations. Therefore, the model is incentivized to allocate its limited attention capacity to truly causal agents to minimize the loss function. 
%In contrast, if the discovered causal graph is misaligned with the true causal interactions, depending on the degree of misalignment, the model would disregard crucial information from actual causal agents leading to a poor performance.

\nbf{Auxiliary Denoising Autoencoder (DAE)} 
Following the definition of the Granger causality for time series data in \secref{sec:method}, the causal graph aids the prediction of future variables from the past value of its parents. Motivated by this we add the DAE task as an auxiliary supervision to facilitate the causal discovery. In this task, the objective is to reconstruct the values of the masked intermediate temporal agent representations generated by AgentNet based on the values of the other nodes and the causal graph. Note that, using temporal features for this task best matches the definition of Granger causality. 

Thereby, we employ a two-layer \textit{graph convolutional network} (GCN) as a \textit{denoising autoencoder} (DAE), where the graph is defined as: $\mathcal{G}_{\text{DAE}}=\{\mathcal{V}_{\text{DAE}}, \mathcal{E}_{\text{DAE}}\}$, the nodes are $\mathcal{V}_{\text{DAE}} = \{v_i^t | v_i^t \in \mathbf{H}^0, i\in\{1,\dots,N\}, t\in\{1,\dots,T'\}\}$, where for computational efficiency we downsample the temporal agent representations, $\mathbf{H}^0 = \text{Flatten}(\text{GroupAvgPool}(\text{SG}(\mathbf{S}'_A))) \in \mathbb{R}^{N_{DAE} \times D_{at}}$, where $N_{\text{DAE}}= T' \times N_{\text{obs}}$, $T'=3$ is the downsampled temporal dimension, and $D_{at}=D_a^1$ is the feature dimension of the temporal agent representations. 
Downsampling is done via group average pooling, with groups defined by chunking the temporal dimension. The tensor \textit{Flatten} operation is then applied so that each node $v_i^t$ corresponds to a specific agent at a particular chunked temporal state. To avoid the model collapse to naïve solutions, we detach the gradients using the \textit{Stop Gradient} operation denoted by $\text{SG}$ so the DAE loss cannot directly affect representation learning in AgentNet.
The edges are defined as $\mathcal{E}_{\text{DAE}} = \{ (v_i^t,v_j^{t'}) | t \leq t', (e_{ij}=1 \; \text{or} \; i=j ) \}$. The edges $\mathcal{E}_{\text{DAE}}$ correspond to the adjacency matrix $\mathbf{A}_{\text{DAE}}$, which is a block lower-triangular extension of the adjacency matrix generated by the CDN.

Next, we mask a random selection of nodes using a binary mask $\mathbf{M} \in \mathbb{R}^{N_{\text{DAE}} \times D_{at}}$ controlled by the masking ratio $r$. The masked representation is given by $\mathbf{H}^1=\mathbf{H}^0 \odot \mathbf{M}$, where $\odot$ is the Hadamard product operator. We constrain the mask to have an all-equal last dimension, i.e., we perform node-wise masking. Subsequently, the GCN layers are defined as: $ \mathbf{H}^2 =\textup{ReLU}\left ( \tilde{\mathbf{A}}_{\text{DAE}}\mathbf{H^1}\mathbf{W^1}\right ), \; \mathbf{H}^3 = \tilde{\mathbf{A}}_{\text{DAE}}\mathbf{H}^2\mathbf{W^2},$
where $\tilde{\mathbf{A}}_{\text{DAE}} \in \mathbb{R}^{N_{\text{DAE}} \times N_{\text{DAE}}}$ is the row-normalized adjacency matrix, $\mathbf{W}^1$ and $\mathbf{W}^2$ are the linear projection weight matrices, and $\mathbf{H}^2$ denotes the intermediate node representations, and $\mathbf{H}^3$ is the reconstructed node representations. Loss function $L_{\textup{DAE}}$ is equal to negative cosine similarity between masked nodes of $\mathbf{H}^0$ and $\mathbf{H}^3$.

\subsection{Causal Trajectory Prediction}
\label{sec:CRiTIC}
\nbf{Causal Attention Gating (CAG)} 
In the Transformer architecture, query tokens attend to the key tokens according to their attention weights controlled by the attention mechanism \cite{vaswani2017attention}.
We propose \textit{causal attention gating} that uses the adjacency matrix of the causal graph generated by CDN to morph the attention weights towards causal agents.
Inspired by \cite{infogate}, we introduce the \textit{CausalAttention} function that is derived by applying CAG to conventional attention mechanism \cite{vaswani2017attention}:
\begin{equation}
\fontsize{9pt}{1pt}\selectfont \textup{CausalAttn}(\mathbf{Q}, \mathbf{K}, \mathbf{V}, \mathbf{A}) = (\mathbf{\Phi} \odot \mathbf{A}) \mathbf{V'} + \alpha(\mathbf{\Phi} \odot \mathbf{A}^C)\; \mathbf{N},
\end{equation}
where $\mathbf{\Phi} = \textup{Softmax}(\frac{\mathbf{Q} \mathbf{K}^T}{\sqrt{d_k}})$,
$A^C = \mathbf{J}-\mathbf{A}$,
$\mathbf{J} \in \mathbb{R}^{N_{obs} \times N_{obs}}$ is an all-ones matrix,
$\mathbf{V}'$ denotes the column-normalized value matrix in attention layers,
$\mathbf{N} = \mathcal{N}(\mathbf{0},\mathbf{1})$ is a Gaussian noise matrix of the shape $\mathbb{R}^{N_{obs} \times d_{v}}$, $d_{v}$ and $d_k$ are the dimensions of the key and value vectors, respectively. The noise scale hyperparameter $\alpha$ is set to zero at the inference time. 

Assuming a satisfactory performance of CDN in estimating causal relations, the model is encouraged to attend to the causal agents as the value of the other agent tokens contains large noise. Therefore their contribution is not productive in lowering the trajectory prediction loss.

\nbf{Backbone Network} 
\label{sec:bacbone}
The backbone network, MTR \cite{mtr}, is composed of 6 layers of Transformer encoder blocks, a dense prediction (DFP) network, and 6 layers of decoder Transformer blocks with latent anchors and dynamic map query layers following \cite{mtr}. In this study we focus on the inter-agent causal discovery, and so unlike the original backbone model, we only feed the map-aware agent representations to the Encoder network and apply CAG for its self-attention layers (with global attention). This modification is only made to isolate and focus on inter-agent causal relationships without interference from map-based interactions. We leave the investigation of agent-map causal interactions as an important direction for future work.

\nbf{Training Objective}
The total loss is a weighted sum of the described loss terms discussed thus far, including \textit{Prediction}, \textit{DFP}, \textit{DAE}, and \textit{Sparsity}, where the weights are tuned as hyperparameters.

\section{Experiments}
\label{sec:experiments}
\subsection{Experimental Settings}
\begin{table}[tp!]
  \centering
  \caption{Robustness results based on RemoveNoncausal perturbation. The metrics are reported for the AV agent, for which we have the causality labels. The upper half of the table shows OnlyAV (OAV) and the lower half PlusAV (PAV) settings. ($\dagger$) indicates results are based on \cite{Causal_agents}, ($\ddagger$)  indicates the model is trained on 20\% of the training set, and for all of the metrics lower values are better. The letter P indicates the metric is reported for the perturbed set and the  S\# indicates sparsity percentage. Letters M and J stand for marginal prediction and joint prediction, respectively.}
  \resizebox{\columnwidth}{!}{
  \begin{tabular}{ l  c @{\hskip 0.2cm}  c @{\hskip 0.2cm} c @{\hskip 0.3cm}  >{\columncolor[gray]{0.8}}c}
    \toprule
    \rowcolor{white!50} {Model} & {\fontsize{6}{8}\selectfont$\text{minADE}$} & {\fontsize{6}{8}\selectfont$\text{minADE}_{P}$} & {\fontsize{6}{8}\selectfont$\Delta\text{minADE}$} & {\fontsize{6}{8}\selectfont$\frac{\Delta\text{minADE}}{ \text{minADE}}$}  \\
    \midrule
    MultiPath++-OAV$\dagger$\cite{multipath++}                           & 0.376 & 0.395 & 0.141  & 37.5\% \\
    SceneTransformer-OAV-M\cite{scene_transformer}       & \textbf{0.250} & \textbf{0.265} & 0.067  & 26.8\% \\
    Wayformer-OAV$\dagger$\cite{wayformer}                               & 0.393 & 0.406 & 0.101 & 25.7\% \\ 
    MTR-OAV-M \cite{mtr}                                         & 0.339 & 0.360 & 0.073  & 21.5\% \\ 
    CRiTIC-OAV-M-SP4 (ours)                                               & 0.389 & 0.399 & \textbf{0.0387}  & \textbf{9.9}  \%  \\
    CRiTIC-OAV-M-SP20 (ours)                                              & 0.362 & 0.380 & 0.0580  & 16.0\%  \\ 
    \hline

    MultiPath++-PAV$\dagger$\cite{multipath++}                      & 0.900 & 0.945 & 0.226  & 25.1\% \\
    SceneTransformer-PAV-J$\dagger$ \cite{scene_transformer}     & 0.493 & 0.504 & 0.170  & 34.5\% \\
    SceneTransformer-PAV-M$\dagger$\cite{scene_transformer}  & \textbf{0.305} & \textbf{0.328} & 0.081  & 26.6\% \\
    MTR-PAV-M$\ddagger$ \cite{mtr}                                     & 0.384 & 0.407 & 0.075 & 19.5\%\\
    CRiTIC-PAV-M-S4$\ddagger$ (ours)                                        & 0.426 & 0.441 & \textbf{0.040 } & \textbf{9.4}\%\\
    CRiTIC-PAV-M-S20$\ddagger$ (ours)                                        & 0.395 & 0.413 & 0.061 & 15.4\%\\
    \bottomrule
  \end{tabular}
  }
  \label{tab:noncausal}
  
\end{table}
\textbf{Dataset:} We use the Waymo Open Motion Dataset (WOMD) \cite{waymo_motion_2021_ICCV} for which 1.1 seconds is used to predict the 8-second future trajectory of target agents.
For the robustness experiments, similar to \cite{Causal_agents}, we modify the set of target agents to have: \textbf{(i) PlusAV} setting for which the autonomous vehicle (AV) (which is specified in the dataset) is added to the set of target agents, and \textbf{(ii) OnlyAV} setting, where the AV is the only target agent.

We also evaluate our model using the INTERACTION \cite{zhan2019interaction} dataset, which contains a diverse set of driving scenarios captured from 11 locations around the globe.
For this dataset, the task is to predict 3 seconds of future trajectory given 1 second of history.

\nbf{Evaluation Metrics} We use the standard benchmark metrics of WOMD and INTERACTION including \textbf{minADE}, \textbf{minFDE}, \textbf{mAP}, and \textbf{Miss Rate (MR)} \cite{waymo_motion_2021_ICCV}.
Following the evaluation protocol in \cite{Causal_agents}, we also assess the model's robustness to causality-based perturbation by reporting \textit{$\Delta$-metrics} given by $\frac{1}{N} \sum_{i=1}^{N} \left | m^i_{\textup{Original}} - m^i_{\textup{Perturbed}}  \right |$.

\nbf{Implementation Details} We set the batch size to 400 for the OnlyAV setting, and 80 for the other settings. We use an AdamW optimizer \cite{loshchilov2018decoupled} with weight decay set to 0.1 to train the model for 30 epochs.
The learning rate is set to $1e-4$, and after epoch 22, it is halved every two epochs. 
A linear learning rate warm-up is also used over the first epoch. 

\subsection{Quantitative Analysis}
\label{exp:qual}
\begin{figure}[tp]
    \centering
    \includegraphics[width=0.99\columnwidth,height=0.2\textheight]{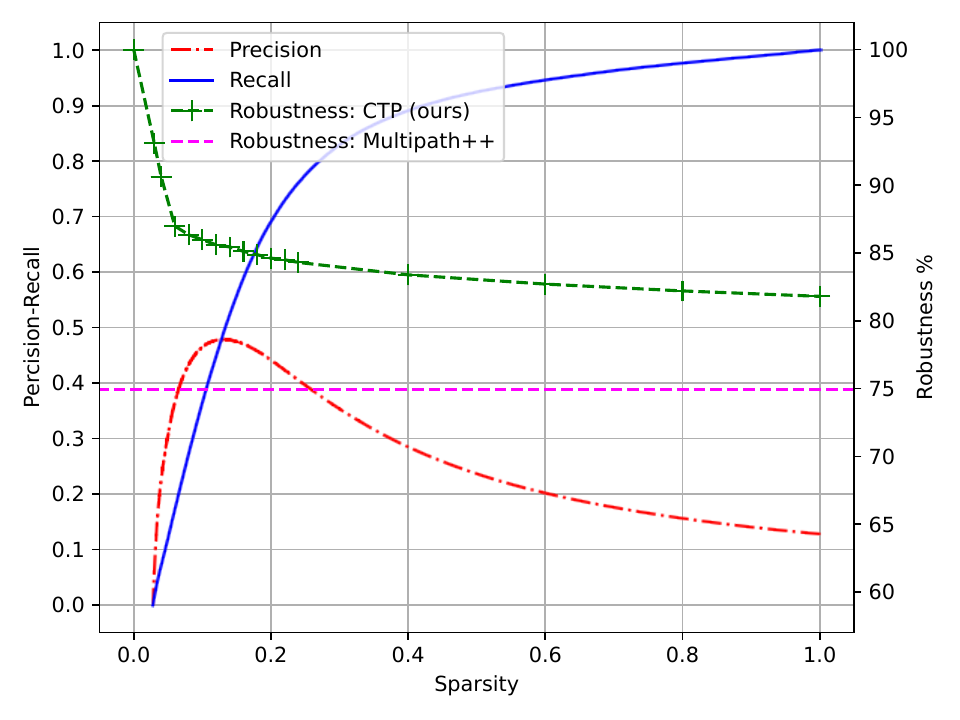}
    \caption{Precision, recall, and the robustness against RemoveNonCausal perturbation for the PlusAV setting. The robustness metric is defined by $(1-\frac{\Delta\text{minADE}}{ \text{minADE}_{\text{Org}}})\times100$.}
    \label{fig:pr_robustness}
\end{figure}

\nbf{Robustness Experiments}
To study the robustness and generalization of the CRiTIC model under causality-based domain shifts, we use scene perturbation tests with the causal agent annotation by Sun et al. \cite{Causal_agents}. In \tabref{tab:noncausal}, we present the result for \textit{RemoveNonCausal} perturbation, where all agents tagged as non-causal are removed.
Evaluating based on the \textit{relative performance drop rate} metric, $\frac{\Delta \textup{minADE}}{\textup{minADE}_{\textup{Original}}}$,
our model shows significant improvement by up to \textbf{54\%} and \textbf{53\%} compared to the next best model for both the OnlyAV and PlusAV settings, respectively.
As expected, the CRiTIC model has less performance drop for RemoveNonCausal perturbations as its attention is directed towards causal agents, and it is less reliant on non-causal agents. 
Note that, we use the minADE metric in the robustness experiments following \cite{Causal_agents}, however, we should note that it is not the primary ranking metric in Waymo's prediction challenge.

Moreover, the sparsity of the predicted causal graph for our model is adjustable at inference time using the threshold hyperparameter.
The sparsity is the ratio of the number of edges in the graph to the number of edges in a fully connected graph (we exclude the self-loop edges in this calculation).
Since we are performing edge classification under the causal structure discovery task, to evaluate the alignment of the predicted edge classes (causal vs. non-causal, which is presented by $e_{ij}$ in \eqref{eq:mpnn}) with the human label \cite{Causal_agents}, we report precision, recall, and robustness under \textit{RemoveNonCausal} perturbation for various causal graph sparsity values (see \figref{fig:pr_robustness}). 

We observe a high correlation between sparsity and robustness and backed by the empirical results, we argue that causal robustness, evaluated by the \textit{RemoveNonCausal} perturbation test, should not be discussed without considering the degree of social attention of the model. For instance, for a socially ignorant (zero sparsity), we can achieve 100\% robustness, but this is not favorable. We advise that the sparsity of the causal graph in the causal prediction model vs. the causal robustness of the model should be considered as a trade-off and the sparsity should be tuned through safety analysis which is beyond the scope of this paper. Thanks to the separate CDN module and an explicit causal interaction graph, we can explicitly measure and control the sparsity of the causal graph for the CRiTIC model. This makes our model more interpretable and more controllable which are critical for the autonomous driving application.

Additionally, we observed that even once the causal gating is removed (when sparsity is 100 \%),  CRiTIC is still more robust than the baselines presented in \tabref{tab:noncausal}. 
This is an indirect influence of CDN in training a causal trajectory model, which persists even if the CDN module is removed. 
We hypothesize that this is due to the information bottleneck, which encourages the model to align its attention weights with the causal graph discovered by CDN.

\nbf{Trajectory Prediction Benchmarking}
In \tabref{tab:waymo_results}, we compare our model with several state-of-the-art models based on the validation and test sets of the WOMD for the marginal motion prediction challenge.
In this challenge, the future motion of the target agents is predicted independently of each other. 
To make a fair comparison, we denoted the models that are evaluated based on ensembling.
Based on these results, we can see that besides achieving better robustness, our model performs similar or better compared to the baseline models.

\begin{table}[tp!]
\centering
  \caption{ Benchmarking results on WOMD. ($\dagger$) indicates result are achieved by using model ensembling, $(\downarrow)$ and $(\uparrow)$ indicate lower and higher values are better.}
\resizebox{\linewidth}{!}{
\begin{tabular}{ll>{\columncolor[gray]{0.8}}cccc}

\toprule 
\rowcolor{white!50}  & Model            & mAP$\uparrow$ & minADE$\downarrow$ & minFDE$\downarrow$ & MissRate$\downarrow$  \\ 
\cmidrule(l){2-2}  \cmidrule(l){3-6}

\multirow{9}{*}{\rotatebox[origin=c]{0}{Test}}  
                      & Wayformer factorized$\dagger$\cite{wayformer} & 0.4120 & 0.5450 & 1.126 & 0.412           \\
                      & Multipath++$\dagger$\cite{multipath++}      &  0.4092 & 0.5557 & 1.1577 & 0.1340           \\
                      & Motion-LM$\dagger$\cite{MotionLM}  & 0.4364 & 0.5509 & 1.1199   & 0.1058  \\ 
                      & MTR-A$\dagger$ \cite{mtr}               & 0.4492 & 0.5640 & 1.1344 &0.1160	          \\  \cdashline{2-6}
                      & MTR \cite{mtr}               & 0.4129 &0.6050 &1.2207 &0.1351	          \\ 
                      & SceneTransformer\cite{scene_transformer} & 0.2790 & 0.6120 & 1.2120        &  0.1560         \\
                      & HDGT\cite{jia2023hdgt}  & 0.2854 &\textbf{0.5933} &\textbf{1.2055}   & 0.1511  \\

                      & CRiTIC (ours)             &      \textbf{0.4174}  &  0.6025    &  1.2276   &   \textbf{0.1342}	       \\ 
\cmidrule(l){1-6}                        
\multirow{3}{*}{\rotatebox[origin=c]{0}{Val.}}  
                      & MTR-A$\dagger$ \cite{mtr}          & 0.4551 & 0.5597 & 1.1299 &0.1167	          \\ \cdashline{2-6}
                      & MTR\cite{mtr}             &  0.4164 & 0.6046 &1.2251&0.1366  \\
                      & CRiTIC (ours)              &   \textbf{0.4190  }&  \textbf{0.5983}    &  \textbf{1.2225}   &   0.1342    \\ \bottomrule
\end{tabular}
} % end resize box
\label{tab:waymo_results}
\end{table}

\nbf{Domain Generalization (DG)} We compare the DG performance of our model with several baselines on various domain splits of the INTERACTION dataset. 
We conduct \textit{cross-scenario} DG experiments (see \cite{li2023cilf} for details) where both the driving scenarios and the locations are different. 
As shown in \tabref{tab:generalization},  CRiTIC shows better performance compared to the baseline models. Based on the results, we hypothesize that the DG improvement is the result of having a causality-aware model. We speculate that part of the domain shift is caused by non-causal agents and since  CRiTIC is empowered by the causal discovery network, it is less dependent on non-causal agents and less influenced by the domain shift created by non-causal agents.
Therefore, being causality-aware leads  CRiTIC to be less affected by domain shift and have better DG performance. 

\subsection{Ablation Study}
We have conducted ablation studies to show the effectiveness of each part in our model.
First, we remove the denoising autoencoder GCN, and then we ablate the whole causal discovery network.
As shown in \tabref{tab:ablation}, we observe that the main robustness improvement against RemoveNonCausal perturbation comes from the MPNN layer. Meanwhile, adding the DAE module has its role in improving the robustness metric and improving the performance of the causal discovery module which is measured by the area under the precision-recall curve (PR-AUC) metric.

Note that forcing the model to only attend to causal agents (by restricting exploitation of spurious features in our case), makes it robust to perturbations but at the cost of justifiable performance drop (less than 1\% mAP drop). 
A non-causal model, on the other hand, has unrestricted access to all of the spuriously correlated features, and the model reaches an slight performance gain by overfitting to them. However, this makes it more vulnerable to the RemoveNonCausal domain shift which breaks the spurious correlations.

\newcommand{\cwidth}{0.12\textwidth}
\begin{table}[tp!]
    \centering
      \caption{ Domain Generalization results on INTERACTION dataset. The ADE, and FDE metrics are shown for test/train domain. The last two columns shows the metric differences across the test and train domains. The results above the dashed line are derived based on \cite{li2023cilf}. ($\dagger$) indicated that the model is augmented based on the CILF method\cite{li2023cilf},  ($\downarrow$) sign indicates the lower value is better. }
      \label{tab:generalization}

\resizebox{\linewidth}{!}{
\begin{tabular}{lcc  >{\columncolor[gray]{0.8}}c  >{\columncolor[gray]{0.8}}c  >{\columncolor[gray]{0.8}}c}
\toprule 
\rowcolor{white!50}  Model & ADE$\downarrow$ & FDE$\downarrow$ & $\Delta$ADE$\downarrow$ & $\Delta$FDE$\downarrow$ \\
\midrule
S-LSTM \cite{alahi2016social} & 2.19/1.13 & 6.62/3.29 & 1.06 & 3.33 \\
CS-LSTM \cite{clstm2018} & 2.22/1.12 & 6.67/3.29 & 1.1 & 3.38 \\
MFP \cite{mfp2019} & 2.17/1.14 & 6.53/3.41 & 1.03 & 3.12 \\
S-LSTM$\dagger$ \cite{li2023cilf} & 2.09/1.07 & 6.37/3.15 & 1.02 & 3.22 \\
CS-LSTM$\dagger$\cite{li2023cilf} & 2.10/1.03 & 6.48/3.09 & 1.07 & 3.39 \\
MFP$\dagger$ \cite{li2023cilf} & 2.10/1.10 & 6.36/3.31 & 1.0 & 3.05 \\
\hdashline
MTR \cite{mtr} & 0.42/0.14 & 1.07/0.34 & 0.28 & 0.73 \\
CRiTIC (Ours) & 0.39/0.19 & 1.07/0.51 & \textbf{0.2} & \textbf{0.56} \\
\bottomrule
\end{tabular}
}
\end{table}

\begin{table}[tp!]
    \centering
      \caption{ Ablation Experiment Results. The study is done in the PlusAV setting. The MPNN is the main body of CDN without DAE and the $\frac{\Delta\text{minADE}}{ \text{minADE}_{\text{Org}}}$ is computed by applying RemoveNonCausal perturbations. $(\downarrow)$ and $(\uparrow)$ indicate lower and higher values are better. }
      \label{tab:ablation}
      \resizebox{\columnwidth}{!}{
    \begin{tabular}{cccccc}
    \toprule
    $\textup{MPNN}$  &   $\text{GCN}_{\text{DAE}}$               & mAP$\uparrow$  & minADE$\downarrow$   & {$\frac{\Delta\text{minADE}}{ \text{minADE}_{\text{Org}}}$} {(\%)$\downarrow$} & PR-AUC $\uparrow$ \\ 
    \cmidrule(l){1-2} \cmidrule(l){3-6} 
    
                  &                    & 0.8238  & 0.3851   &   23.6 \%   & NA  \\
     \checkmark   &                    & 0.8177  & 0.3910   &   15.7 \%  &  0.348 \\
     \checkmark   &   \checkmark       & 0.8179  & 0.3951   &   15.4 \%  &  0.373 \\ \hline
    \end{tabular}
    }
\end{table}

\section{Conclusion}
We proposed a novel causal trajectory prediction model, namely CRiTIC, that is significantly more robust against causal perturbations compared to state-of-the-art trajectory prediction models.
Our framework utilizes a causal discovery network in which we incorporated a self-supervised denoising auxiliary task to guide the causal discovery.
Moreover, we proposed a novel causal attention-gating mechanism that when combined with sparsity regularization of the causal graph, creates an information bottleneck and directs the model's limited attention capacity in the Transformer architecture toward the causal agents. 

We conducted extensive experimentation, including agent removal perturbation tests on our model and state-of-the-art prediction models. The results showed our model achieves up to $\mathbf{54\%}$ improvement in terms of a causality-based robustness metric with minimal degradation in the performance.
In addition, by conducting single-scenario and cross-scenario domain generalization experiments, we observed a better DG performance of the CRiTIC compared with the baseline models. In the future, we intend to extend our approach to address the issue of confounder variables and  explicitly take into account road elements in the causal graph.

\label{sec:conclusion}

\bibliographystyle{ieeetr}
\bibliography{ms}

\begin{thebibliography}{10}

\bibitem{multipath++}
B.~Varadarajan, A.~Hefny, A.~Srivastava, K.~S. Refaat, N.~Nayakanti,
  A.~Cornman, K.~Chen, B.~Douillard, C.~P. Lam, D.~Anguelov, {\em et~al.},
  ``{Multipath++}: Efficient information fusion and trajectory aggregation for
  behavior prediction,'' in {\em ICRA}, 2022.

\bibitem{mtr}
S.~Shi, L.~Jiang, D.~Dai, and B.~Schiele, ``Motion transformer with global
  intention localization and local movement refinement,'' in {\em NeurIPS},
  2022.

\bibitem{scene_transformer}
J.~Ngiam, V.~Vasudevan, B.~Caine, Z.~Zhang, H.~T.~L. Chiang, J.~Ling,
  R.~Roelofs, A.~Bewley, C.~Liu, A.~Venugopal, D.~J. Weiss, B.~Sapp, Z.~Chen,
  and J.~Shlens, ``Scene transformer: A unified architecture for predicting
  future trajectories of multiple agents,'' in {\em ICLR}, 2022.

\bibitem{wayformer}
N.~Nayakanti, R.~Al~Rfou, A.~Zhou, K.~Goel, K.~S. Refaat, and B.~Sapp,
  ``Wayformer: Motion forecasting via simple \& efficient attention networks,''
  in {\em ICRA}, 2023.

\bibitem{Causal_agents}
L.~Sun, R.~Roelofs, B.~Caine, K.~S. Refaat, B.~Sapp, S.~Ettinger, and W.~Chai,
  ``{CausalAgents}: A robustness benchmark for motion forecasting,'' in {\em
  ICRA}, 2024.

\bibitem{causal_representation_learning}
B.~Schölkopf, F.~Locatello, S.~Bauer, N.~R. Ke, N.~Kalchbrenner, A.~Goyal, and
  Y.~Bengio, ``Toward causal representation learning,'' {\em Proceedings of the
  IEEE}, 2021.

\bibitem{Causal_im}
M.~R. Samsami, M.~Bahari, S.~Salehkaleybar, and A.~Alahi, ``Causal imitative
  model for autonomous driving,'' {\em arXiv:2112.03908}, 2021.

\bibitem{liu2022towards}
Y.~Liu, R.~Cadei, J.~Schweizer, S.~Bahmani, and A.~Alahi, ``Towards robust and
  adaptive motion forecasting: A causal representation perspective,'' in {\em
  CVPR}, 2022.

\bibitem{zhu2021survey}
Y.~Zhu, W.~Xu, J.~Zhang, Y.~Du, J.~Zhang, Q.~Liu, C.~Yang, and S.~Wu, ``A
  survey on graph structure learning: Progress and opportunities,'' {\em
  arXiv:2103.03036}, 2021.

\bibitem{NRI_18}
T.~Kipf, E.~Fetaya, K.-C. Wang, M.~Welling, and R.~Zemel, ``Neural relational
  inference for interacting systems,'' in {\em ICML}, 2018.

\bibitem{IDGL}
Y.~Chen, L.~Wu, and M.~Zaki, ``Iterative deep graph learning for graph neural
  networks: Better and robust node embeddings,'' in {\em NeurIPS}, 2020.

\bibitem{franceschi2019learning}
L.~Franceschi, M.~Niepert, M.~Pontil, and X.~He, ``Learning discrete structures
  for graph neural networks,'' in {\em ICML}, 2019.

\bibitem{SLAPS_lgi21}
B.~Fatemi, L.~E. Asri, and S.~M. Kazemi, ``{SLAPS}: Self-supervision improves
  structure learning for graph neural networks,'' in {\em NeurIPS}, 2021.

\bibitem{entner2010causal}
D.~Entner and P.~O. Hoyer, ``On causal discovery from time series data using
  {FCI},'' {\em Probabilistic Graphical Models}, pp.~121--128, 2010.

\bibitem{chickering2002optimal}
D.~M. Chickering, ``Optimal structure identification with greedy search,'' {\em
  JMLR}, vol.~3, pp.~507--554, 2002.

\bibitem{tank2021neural}
A.~Tank, I.~Covert, N.~Foti, A.~Shojaie, and E.~B. Fox, ``Neural {Granger}
  causality,'' {\em PAMI}, vol.~44, no.~8, pp.~4267--4279, 2021.

\bibitem{amortized_causal_discovery}
S.~L{\"o}we, D.~Madras, R.~Zemel, and M.~Welling, ``Amortized causal discovery:
  Learning to infer causal graphs from time-series data,'' in {\em CLeaR},
  2022.

\bibitem{Rasouli_ICRA_2023}
A.~Rasouli and I.~Kotseruba, ``Pedformer: Pedestrian behavior prediction via
  cross-modal attention modulation and gated multitask learning,'' in {\em
  ICRA}, 2023.

\bibitem{li2022graph}
L.~Li, M.~Pagnucco, and Y.~Song, ``Graph-based spatial transformer with memory
  replay for multi-future pedestrian trajectory prediction,'' in {\em CVPR},
  2022.

\bibitem{Bae_2022_ECCV}
I.~Bae, J.-H. Park, and H.-G. Jeon, ``Learning pedestrian group representations
  for multi-modal trajectory prediction,'' in {\em ECCV}, 2022.

\bibitem{Choi_2024_IV}
Y.~Choi, R.~C. Mercurius, S.~Mohamad Alizadeh~Shabestary, and A.~Rasouli,
  ``Dice: Diverse diffusion model with scoring for trajectory prediction,'' in
  {\em IV}, 2024.

\bibitem{rasouli2021bifold}
A.~Rasouli, M.~Rohani, and J.~Luo, ``Bifold and semantic reasoning for
  pedestrian behavior prediction,'' in {\em ICCV}, 2021.

\bibitem{Shi_2021_CVPR}
L.~Shi, L.~Wang, C.~Long, S.~Zhou, M.~Zhou, Z.~Niu, and G.~Hua, ``{SGCN}:
  Sparse graph convolution network for pedestrian trajectory prediction,'' in
  {\em CVPR}, 2021.

\bibitem{Pourkeshavarz_2024_CVPR}
M.~Pourkeshavarz, J.~Zhang, and A.~Rasouli, ``Cadet: a causal disentanglement
  approach for robust trajectory prediction in autonomous driving,'' in {\em
  CVPR}, 2024.

\bibitem{Karim_2024_ICRA}
R.~Karim, S.~M.~A. Shabestary, and A.~Rasouli, ``Destine: Dynamic goal queries
  with temporal transductive alignment for trajectory prediction,'' in {\em
  ICRA}, 2024.

\bibitem{kim2021lapred}
B.~Kim, S.~H. Park, S.~Lee, E.~Khoshimjonov, D.~Kum, J.~Kim, J.~S. Kim, and
  J.~W. Choi, ``{LaPred}: Lane-aware prediction of multi-modal future
  trajectories of dynamic agents,'' in {\em CVPR}, 2021.

\bibitem{zhou2022hivt}
Z.~Zhou, L.~Ye, J.~Wang, K.~Wu, and K.~Lu, ``{HiVT}: Hierarchical vector
  transformer for multi-agent motion prediction,'' in {\em CVPR}, 2022.

\bibitem{wang2022ltp}
J.~Wang, T.~Ye, Z.~Gu, and J.~Chen, ``{LTP}: Lane-based trajectory prediction
  for autonomous driving,'' in {\em CVPR}, 2022.

\bibitem{liang2020learning}
M.~Liang, B.~Yang, R.~Hu, Y.~Chen, R.~Liao, S.~Feng, and R.~Urtasun, ``Learning
  lane graph representations for motion forecasting,'' in {\em ECCV}, 2020.

\bibitem{amirloo2022latentformer}
E.~Amirloo, A.~Rasouli, P.~Lakner, M.~Rohani, and J.~Luo, ``{LatentFormer}:
  Multi-agent transformer-based interaction modeling and trajectory
  prediction,'' {\em arXiv:2203.01880}, 2022.

\bibitem{Cui_2021_ICCV}
A.~Cui, S.~Casas, A.~Sadat, R.~Liao, and R.~Urtasun, ``{LookOut}: Diverse
  multi-future prediction and planning for self-driving,'' in {\em ICCV}, 2021.

\bibitem{Casas_2020_ECCV}
S.~Casas, C.~Gulino, S.~Suo, K.~Luo, R.~Liao, and R.~Urtasun, ``Implicit latent
  variable model for scene-consistent motion forecasting,'' in {\em ECCV},
  2020.

\bibitem{Girgis_2022_ICLR}
R.~Girgis, F.~Golemo, F.~Codevilla, M.~Weiss, J.~A. D'Souza, S.~E. Kahou,
  F.~Heide, and C.~Pal, ``Latent variable sequential set transformers for joint
  multi-agent motion prediction,'' in {\em ICLR}, 2022.

\bibitem{gao2020vectornet}
J.~Gao, C.~Sun, H.~Zhao, Y.~Shen, D.~Anguelov, C.~Li, and C.~Schmid,
  ``{VectorNet}: Encoding hd maps and agent dynamics from vectorized
  representation,'' in {\em CVPR}, 2020.

\bibitem{Pourkeshavarz_2023_ICCV}
M.~Pourkeshavarz, C.~Chen, and A.~Rasouli, ``Learn tarot with mentor: A
  meta-learned self-supervised approach for trajectory prediction,'' in {\em
  ICCV}, 2023.

\bibitem{Zhang_2024_IV}
J.~Zhang, M.~Pourkeshavarz, and A.~Rasouli, ``Tract: A training dynamics aware
  contrastive learning framework for long-tail trajectory prediction,'' in {\em
  IV}, 2024.

\bibitem{mfp2019}
C.~Tang and R.~R. Salakhutdinov, ``Multiple futures prediction,'' in {\em
  NeurIPS}, 2019.

\bibitem{Salzmann_2020_ECCV}
T.~Salzmann, B.~Ivanovic, P.~Chakravarty, and M.~Pavone, ``Trajectron++:
  Multi-agent generative trajectory forecasting with heterogeneous data for
  control,'' in {\em ECCV}, 2020.

\bibitem{gilles2022gohome}
T.~Gilles, S.~Sabatini, D.~Tsishkou, B.~Stanciulescu, and F.~Moutarde,
  ``{GOHOME}: Graph-oriented heatmap output for future motion estimation,'' in
  {\em ICRA}, 2022.

\bibitem{lee2022muse}
M.~Lee, S.~S. Sohn, S.~Moon, S.~Yoon, M.~Kapadia, and V.~Pavlovic,
  ``{MUSE-VAE}: Multi-scale {VAE} for environment-aware long term trajectory
  prediction,'' in {\em CVPR}, 2022.

\bibitem{zhao2021tnt}
H.~Zhao, J.~Gao, T.~Lan, C.~Sun, B.~Sapp, B.~Varadarajan, Y.~Shen, Y.~Shen,
  Y.~Chai, C.~Schmid, {\em et~al.}, ``{TNT}: Target-driven trajectory
  prediction,'' in {\em CoRL}, 2020.

\bibitem{Rasouli_2024_ICRA}
A.~Rasouli, ``A novel benchmarking paradigm and a scale- and motion-aware model
  for egocentric pedestrian trajectory prediction,'' in {\em ICRA}, 2024.

\bibitem{yuan2021agentformer}
Y.~Yuan, X.~Weng, Y.~Ou, and K.~M. Kitani, ``{AgentFormer}: Agent-aware
  transformers for socio-temporal multi-agent forecasting,'' in {\em ICCV},
  2021.

\bibitem{clstm2018}
N.~Deo and M.~M. Trivedi, ``Convolutional social pooling for vehicle trajectory
  prediction,'' in {\em CVPRW}, 2018.

\bibitem{chen2021human}
G.~Chen, J.~Li, J.~Lu, and J.~Zhou, ``Human trajectory prediction via
  counterfactual analysis,'' in {\em ICCV}, 2021.

\bibitem{granger1969investigating}
C.~W. Granger, ``Investigating causal relations by econometric models and
  cross-spectral methods,'' {\em Econometrica: Journal of the Econometric
  Society}, pp.~424--438, 1969.

\bibitem{pointnet}
C.~R. Qi, H.~Su, K.~Mo, and L.~J. Guibas, ``{PointNet}: Deep learning on point
  sets for 3d classification and segmentation,'' in {\em CVPR}, 2017.

\bibitem{gru}
R.~Socher, C.~D. Manning, and A.~Y. Ng, ``Learning continuous phrase
  representations and syntactic parsing with recursive neural networks,'' in
  {\em NeurIPS}, 2010.

\bibitem{gilmer2017mpnn}
J.~Gilmer, S.~S. Schoenholz, P.~F. Riley, O.~Vinyals, and G.~E. Dahl, ``Neural
  message passing for quantum chemistry,'' in {\em ICML}, 2017.

\bibitem{gumbel_trick}
C.~J. Maddison, A.~Mnih, and Y.~W. Teh, ``The concrete distribution: A
  continuous relaxation of discrete random variables,'' in {\em ICLR}, 2017.

\bibitem{vaswani2017attention}
A.~Vaswani, N.~Shazeer, N.~Parmar, J.~Uszkoreit, L.~Jones, and Others,
  ``Attention is all you need,'' in {\em NeurIPS}, 2017.

\bibitem{infogate}
M.~Tomar, R.~Islam, M.~E. Taylor, S.~Levine, and P.~Bachman, ``Ignorance is
  bliss: Robust control via information gating,'' in {\em NeurIPS}, 2023.

\bibitem{waymo_motion_2021_ICCV}
S.~Ettinger, S.~Cheng, B.~Caine, C.~Liu, H.~Zhao, S.~Pradhan, Y.~Chai, B.~Sapp,
  C.~R. Qi, Y.~Zhou, Z.~Yang, A.~Chouard, P.~Sun, J.~Ngiam, V.~Vasudevan,
  A.~McCauley, J.~Shlens, and D.~Anguelov, ``Large scale interactive motion
  forecasting for autonomous driving: The waymo open motion dataset,'' in {\em
  ICCV}, 2021.

\bibitem{zhan2019interaction}
W.~Zhan, L.~Sun, D.~Wang, H.~Shi, A.~Clausse, M.~Naumann, J.~Kummerle,
  H.~Konigshof, C.~Stiller, A.~de~La~Fortelle, {\em et~al.}, ``{INTERACTION}
  dataset: An international, adversarial and cooperative motion dataset in
  interactive driving scenarios with semantic maps,'' {\em arXiv:1910.03088},
  2019.

\bibitem{loshchilov2018decoupled}
I.~Loshchilov and F.~Hutter, ``Decoupled weight decay regularization,'' in {\em
  ICLR}, 2019.

\bibitem{MotionLM}
A.~Seff, B.~Cera, D.~Chen, M.~Ng, A.~Zhou, N.~Nayakanti, K.~S. Refaat,
  R.~Al-Rfou, and B.~Sapp, ``{MotionLM}: Multi-agent motion forecasting as
  language modeling,'' in {\em ICCV}, 2023.

\bibitem{jia2023hdgt}
X.~Jia, P.~Wu, L.~Chen, Y.~Liu, H.~Li, and J.~Yan, ``{HDGT}: Heterogeneous
  driving graph transformer for multi-agent trajectory prediction via scene
  encoding,'' {\em PAMI}, 2023.

\bibitem{li2023cilf}
S.~Li, Q.~Xue, Y.~Zhang, and X.~Li, ``{CILF}: Causality inspired learning
  framework for out-of-distribution vehicle trajectory prediction,'' in {\em
  Asian Conference on Pattern Recognition}, 2023.

\bibitem{alahi2016social}
A.~Alahi, K.~Goel, V.~Ramanathan, A.~Robicquet, L.~Fei-Fei, and S.~Savarese,
  ``Social {LSTM}: Human trajectory prediction in crowded spaces,'' in {\em
  CVPR}, 2016.

\end{thebibliography}

\end{document}